\documentclass[10pt,twocolumn,letterpaper]{article}

\usepackage{cvpr} 

\usepackage{CJK}
\usepackage{multicol}
\usepackage{color}
\usepackage{xcolor}
\usepackage{colortbl}
\usepackage{bm}
\usepackage{extarrows}
\usepackage{float}
\usepackage{bbm}
\usepackage{booktabs}
\usepackage{multirow}
\usepackage{amsmath}
\usepackage{amssymb}
\usepackage{algorithm}
\usepackage{algorithmic}
\usepackage{makecell}
\usepackage{amsmath}
\usepackage{amssymb}
\usepackage{amsthm}
\usepackage[utf8]{inputenc}
\usepackage{amsfonts} 
\usepackage{caption} 
\usepackage{newfloat}
\usepackage{listings}

\usepackage{xcolor}
\definecolor{thmcolor}{HTML}{0201F5}

\newtheoremstyle{coloredthm}
  {3pt}
  {3pt}
  {\itshape}
  {}
  {\bfseries\color{thmcolor}}
  {.}
  {.5em}
  {}

\theoremstyle{coloredthm}

\definecolor{cvprblue}{rgb}{0.21,0.49,0.74}
\definecolor{supplcolor}{HTML}{A32D26}

\definecolor{noise20color}{HTML}{FFF9E6} 
\definecolor{noise50color}{HTML}{EDF7ED} 
\definecolor{noise80color}{HTML}{EBF5FB} 

\usepackage[pagebackref,breaklinks,colorlinks,citecolor=brown]{hyperref}
\usepackage{thmtools}

\def\paperID{*****} 
\def\confName{CVPR}
\def\confYear{2026}

\def\modelname{\mbox{EgoAdapt} }

\newcommand\blfootnote[1]{%
  \begingroup
  \renewcommand\thefootnote{}\footnote{#1}%
  \addtocounter{footnote}{-1}%
  \endgroup
}

\title{EgoAdapt: A Multi-Scene Egocentric Adaptation Method for CVPR 2026 HD-EPIC VQA Challenge}

\author{Zhiwei Chen$^{1}$~~~~Yupeng Hu$^{1}$\blfootnote{~corresponding authors}~~~~Zixu Li$^{1}$~~~~Zhiheng Fu$^{1}$~~~~Guozhi Qiu$^{1}$~~~~Weili Guan$^{2}$~~~~Liqiang Nie$^{2}$ \vspace{2mm}\\
$^1$Shandong University\hspace{1.5cm}$^2$Harbin Institute of Technology (Shenzhen)\hspace{1.5cm}\\
{\tt\small \{zivczw, lizixu.cs, fuzhiheng8, gzqiu007, honeyguan, nieliqiang\}@gmail.com;} \\ 
{\tt\small \ 
huyupeng@sdu.edu.cn}
}
\begin{document}

\newcommand{\method}{EgoAdapt}
\newcommand{\dataset}{HD-EPIC}
\newcommand{\qwen}{Qwen3-VL-8B}

\maketitle
\begin{abstract}
This technical report presents our solution, \modelname (Egocentric Adaptation via Category, Calibration, and Consistency), to the CVPR 2026 HD-EPIC VQA challenge.
HD-EPIC evaluates whether a vision-language model can reason over realistic first-person kitchen videos, where the evidence for an answer may be a short hand-object interaction, a long recipe trajectory, a spatial relation to a fixture, or a subtle gaze cue.
The benchmark contains 26K multiple-choice questions across seven macro-categories: recipe, ingredient, nutrition, fine-grained action, 3D perception, object motion, and gaze.
We observe that the main difficulty is not only model capacity, but also the mismatch between a single generic inference recipe and the heterogeneous temporal, spatial, and semantic structure of the benchmark.
Our method, \method, introduces three inference-time components: (1) category-conditioned routing with per-category prompts, frame budgets, and sampling rates; (2) calibrated option scoring that evaluates all candidate answers with letter-token likelihoods and generation agreement instead of relying only on direct generation; and (3) test-time consistency adaptation that aggregates predictions across option permutations and verification-style prompts for ambiguous cases.
This design substantially improves over the available HD-EPIC baselines.
\end{abstract}

\section{Introduction}
\label{sec:intro}

Egocentric video understanding~\cite{perrett2025hd,TempRet} has moved beyond recognizing isolated actions: a useful model must connect hand-object interactions, scene geometry, object state changes, and human intent over long, untrimmed recordings.
This requirement is especially visible in kitchens, where visually similar motions can serve different goals, and where the relevant evidence may be distributed across many moments of activity.
More broadly, recent progress in language-conditioned visual retrieval has shown that fine-grained visual understanding often depends ondisentangling entities, modifications, and relational cues rather than matching global semantics alone~\cite{thawakar2026covr,MEDIAN,FineCIR,STABLE}.
HD-EPIC~\cite{perrett2025hd} provides a highly-detailed benchmark for this setting.
It contains 26K multiple-choice VQA questions organized under seven macro-categories: recipe, ingredient, nutrition, fine-grained action, 3D perception, object motion, and gaze.
These categories collectively probe a broad range of egocentric abilities, from recipe step localization and ingredient tracking to spatial grounding, object itinerary reconstruction, and interaction anticipation.

The benchmark is challenging for current video-language models for three reasons.
First, the temporal scale varies dramatically across questions.
A fine-grained action query may depend on a brief manipulation lasting less than a second, while a recipe or object-motion query may require preserving context~\cite{ReTrack,li2023uniformerv2,HUD,ConeSep,kuang2021video,li2024mvbench,ERASE} over several moments or across multiple input videos.
Second, many questions contain explicit temporal or spatial cues, which require the model to connect language with the relevant visual evidence instead of answering from global video context alone. Similar challenges have also been studied in composed video retrieval and composed image retrieval~\cite{HABIT,Air-Know,MELT}, where models must preserve shared semantics while identifying discriminative temporal or directional evidence~\cite{ENCODER,ventura2024covr,OFFSET,PAIR,HINT,INTENT,REFINE,TEMA}.
Third, the seven macro-categories do not share the same reasoning structure.
Nutrition questions require commonsense comparison of ingredients and changes over time; 3D perception questions require spatial grounding; and recipe questions require matching observed procedures to a latent cooking plan.
This heterogeneity makes a single prompt a poor fit for the entire benchmark.

Our work treats these limitations as an alignment problem.
The central design principle is simple: the model should receive the temporal resolution, visual coverage, and semantic framing that best match the evidence required by that question~\cite{zhang2024llava,cheng2024videollama}.
We therefore route each question through a category-specific configuration, score every candidate answer explicitly, and invoke consistency checks only when the model's option scores indicate ambiguity.

For HD-EPIC, this structure is unusually important.
The same video may support questions about what recipe is being cooked, which ingredient is being added, where a fixture is located, and what object will be touched next.
Each of these questions asks the model to attend to a different slice of the same egocentric stream.
Our category-conditioned test-time adaptation makes this difference explicit while preserving a unified multiple-choice interface.

Our contributions are:
\begin{itemize}
    \item We present a systematic pipeline, \method, for egocentric VQA that improves HD-EPIC performance.
    \item We introduce category-conditioned inference routing with per-category prompts tailored to the semantic and temporal structure of each macro-category.
    \item We use calibrated option scoring with generation agreement and consistency-based disambiguation, reducing the instability of direct letter generation in multiple-choice video QA.
    \item We provide an analysis of the resulting gains on the HD-EPIC challenge, offering practical guidance for future methods on heterogeneous egocentric benchmarks.
\end{itemize}

\section{Method}
\label{sec:method}

\modelname is designed as a question-aware inference framework for HD-EPIC rather than as a single prompt template.
The core idea is to decompose the multiple-choice decision into three controllable stages: selecting the appropriate video evidence, converting candidate answers into comparable model scores, and stabilizing uncertain decisions through consistency checks.
For each HD-EPIC question, \modelname first converts the raw benchmark instance into a multiple-choice prompt, then selects a category-specific inference route, scores all answer options, and finally applies consistency-based adaptation when the score distribution is ambiguous.
This design keeps the backbone interface simple while giving the pipeline enough structure to handle the different question families in HD-EPIC.

\begin{figure*}[t]
\centering
\includegraphics[width=\textwidth]{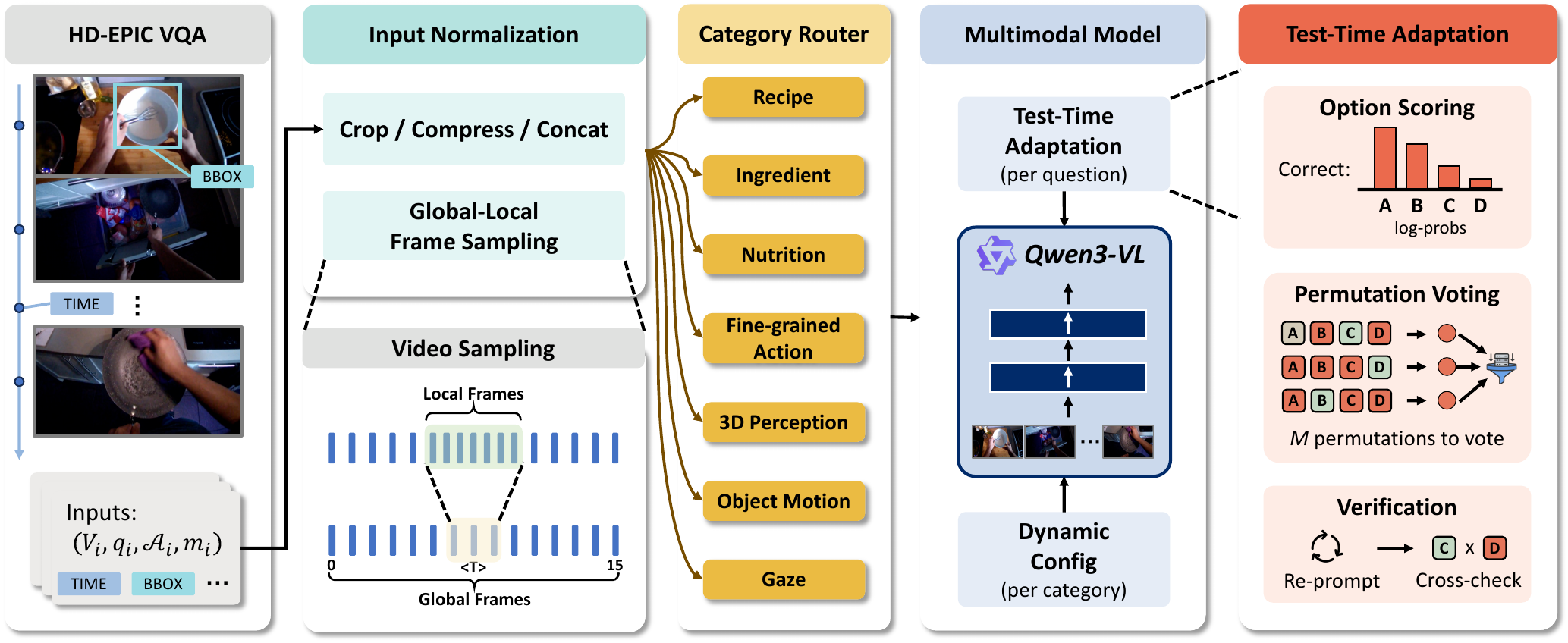}
\caption{Pipeline of \method. It formats each benchmark question as multiple-choice VQA, routes it to a category-specific configuration, scores all answer options with log-probabilities, and applies consistency adaptation for ambiguous cases.}
\label{fig:pipeline}
\end{figure*}

\subsection{Category-Conditioned Routing}
\label{sec:routing}

The HD-EPIC benchmark spans seven macro-categories with fundamentally different visual and temporal characteristics.
Recipe questions require long-horizon context to localize preparation steps and distinguish interleaved recipes.
Fine-grained action questions demand dense observation of short hand motions.
Gaze questions depend heavily on head direction and hand trajectory in the final seconds, whereas 3D perception and object-motion questions often require tracking objects and fixtures through longer intervals.
Using a single inference configuration for all questions forces an unnecessary compromise between temporal coverage and temporal resolution.

In \method, we route each question through a category-specific configuration.
The route controls three elements: the system prompt, the sampling FPS, and the maximum frame budget $N_c$.
Given a segment of length $T$, frames are sampled uniformly under the category frame budget:
\begin{equation}
\begin{aligned}
    \mathcal{F} &= \{f(t_1), \ldots, f(t_N)\}, \\
    t_j &= t_{\mathrm{start}} + \frac{j-1}{N-1}(t_{\mathrm{end}}-t_{\mathrm{start}}),
\end{aligned}
\end{equation}
with $N \leq N_c$.
Long-context categories such as recipe, ingredient, nutrition, 3D perception, and object motion are assigned larger frame budgets so that the model can preserve event order over minutes.
Short-window categories such as fine-grained action and gaze use higher effective temporal density within a smaller window.
This routing step is intentionally coarse-grained.
\modelname does not attempt to design a separate solver for every sub-task; instead, it groups questions by the dominant type of evidence they require.
This choice gives the system enough specialization to avoid an unsuitable generic inference setting, while retaining a shared implementation across all macro-categories.

\textbf{Evidence allocation.}
The frame budget in \modelname is treated as an evidence allocation problem.
For recipe, ingredient, nutrition, 3D perception, and object motion, missing a key event can make the answer impossible, so \modelname prioritizes broad temporal coverage.
For fine-grained action and gaze, the discriminative signal is usually local and transient, so \modelname increases the effective temporal density within a shorter observation window.
This difference is particularly important for egocentric video: a single frame may reveal the object being manipulated, but the action intent often depends on the preceding or following motion.

\textbf{Answer options with video segments.}
Some HD-EPIC questions present answer candidates that are themselves temporal references within the input video(s), expressed as either single timestamps or intervals (e.g., ``video\,1 from 00:08:00 to 00:08:01'').
In the benchmark annotations, these intervals are encoded via \texttt{<TIME>} tags that reference the original video timestamps and input keys.
An answer option may consist of a single interval (as in fine-grained action localization) or multiple discontinuous intervals spanning one or two input videos (as in recipe step localization).
In \method, all answer options are rendered as text labels within the multiple-choice prompt; no separate per-option video segments are extracted.
Before being presented to the model, every \texttt{<TIME>} tag in both the question and the answer choices is remapped to the temporal coordinates of the subsampled video that the model actually observes:
\begin{equation}
    t' = \frac{t_{\mathrm{raw}} - t_{\mathrm{input\_start}}}{D} + o_k,
\end{equation}
where $D$ is the category-specific temporal divisor and $o_k$ is the concatenation offset for multi-video inputs.
This remapping ensures that the time labels in the answer text are directly aligned with the frame indices visible to the model, allowing it to localize the relevant sub-intervals within the single category-conditioned video input.

\textbf{System prompt design.}
The system prompt in \modelname injects task structure that is already present in the benchmark taxonomy.
For recipe questions, the prompt emphasizes step localization, step recognition, multi-recipe tracking, and preparation-step identification.
Ingredient questions focus on scale reading, quantity estimation, ingredient recognition, and addition order.
Nutrition questions encourage comparison of carb, fat, protein, and calorie contributions and ask the model to track nutritional changes when ingredients are added.
Fine-grained action questions highlight action recognition, temporal localization, and how/why reasoning.
3D perception questions focus on bounding-box grounding, fixture interaction counting, spatial direction, object placement, and contents retrieval.
Object-motion questions ask the model to count location changes, reconstruct movement itineraries, and identify stationary periods.
Gaze questions separate current gaze estimation from future interaction anticipation.
The prompt format remains fixed; only the semantic instruction changes with the route.
This keeps the downstream scoring interface identical across categories.
As a result, the router changes what the model should attend to and how much video it receives, but the final answer selection remains a unified multiple-choice decision.

\subsection{Calibrated Option Scoring}
\label{sec:option_scoring}

The second component of \modelname is calibrated option scoring.
Direct letter generation is brittle in multiple-choice video QA.
Even when a model understands the video, it may output an unconstrained phrase, include extra explanation, or switch answers under small changes in option order.
Within \method, we therefore adopt a score-then-select strategy.
Instead of asking the model to only generate a letter, we read the model's next-token distribution over option letters and select the option with the highest calibrated score.

For a prompt $P_i$ ending with \texttt{Correct:}, the model assigns a probability to each option letter.
We define the base score of option $k$ as the log-probability of its letter token $\ell_k$ at the first generation position:
\begin{equation}
    s_{i,k} = \log p_\theta(\ell_k \mid \mathcal{F}_i, P_i).
    \label{eq:option_score}
\end{equation}
This requires one forward pass through the model, and the scores for all options are obtained from the same output logits.
Because every candidate answer is evaluated under the same prompt and visual input, the resulting scores are directly comparable.
This is useful for HD-EPIC because many answer options are semantically close: two time intervals may overlap in the same recipe stage, two object locations may be visually adjacent, or two gaze targets may appear in the same region of the frame.
In such cases, relying only on the text generated by the model hides useful uncertainty, whereas \modelname exposes the full option score vector.
In parallel, we run a short free-generation pass and parse the generated answer $\hat{z}$ into an option index $\hat{k}$ when possible.
The final score combines likelihood and generation agreement:
\begin{equation}
    \tilde{s}_{i,k} = s_{i,k} + \beta\,\mathbf{1}[k=\hat{k}],
    \label{eq:final_score}
\end{equation}
where $\beta$ is a small agreement bonus.
This design keeps the robustness of likelihood-based selection while still using the model's direct generation as a weak confirmation signal.
If the free-generation answer cannot be parsed as a valid option, \modelname falls back to the likelihood scores alone.
Thus, invalid or verbose generations do not become invalid submissions, and the score vector remains available for confidence estimation.
The top-score margin is later used by the consistency module to decide whether additional inference is warranted.

\subsection{Test-Time Consistency Adaptation}
\label{sec:tta}

The final component of \modelname addresses residual ambiguity.
Some HD-EPIC questions remain ambiguous even after category routing and calibrated scoring.
Typical cases include visually similar recipe steps, object-motion options that differ only in a small temporal interval, or gaze questions where several candidate objects lie close to the center of view.
For these questions, the gap between the top two option scores is small.
\modelname uses this score gap as a trigger for consistency adaptation, avoiding unnecessary extra inference on clear cases.

\textbf{Option permutation voting.}
For triggered questions, \modelname re-runs inference $M$ times with the answer options presented in different random orders $\pi_m$.
Each run produces a predicted option $\hat{y}^{(m)}$ in the shuffled order.
We map it back to the original option index and take a majority vote:
\begin{equation}
    \hat{y} = \operatorname{argmax}_k \sum_{m=1}^{M}\mathbf{1}[\pi_m^{-1}(\hat{y}^{(m)})=k].
    \label{eq:vote}
\end{equation}
This test probes whether the model's evidence is tied to the answer content rather than to the presentation position.
A stable prediction should survive option shuffling, whereas a weak or position-biased prediction is more likely to flip.
The vote therefore acts as a practical calibration mechanism.
It does not assume that every shuffled run is correct; instead, it checks whether the same underlying option repeatedly emerges when superficial ordering cues are removed.
This is especially helpful for multiple-choice benchmarks where option position can become an unintended shortcut.

\textbf{Verification re-prompting.}
For categories where a compact auxiliary question can summarize the evidence, such as nutrition and object motion, we additionally issue a simplified verification prompt.
The verification prompt asks for an intermediate fact, for example which ingredient was added or whether an object changed its resting location.
In our implementation this signal is used conservatively: it provides an additional consistency check around the primary multiple-choice prediction rather than replacing the full question-specific decision rule.
This preserves the benchmark's answer format while allowing the pipeline to inspect whether the model's intermediate observation is compatible with the selected option.
Together, permutation voting and verification re-prompting make \modelname a confidence-aware inference system: easy questions are answered with one calibrated pass, while hard questions receive additional consistency checks.

\section{Experiments}
\label{sec:experiments}

\begin{table*}[t]
\centering
\small
\begin{tabular}{@{}lccccccc|c@{}}
\Xhline{1pt}
Method & \rotatebox{60}{Recipe} & \rotatebox{60}{Ingr.} & \rotatebox{60}{Nutr.} & \rotatebox{60}{Fine-gr.} & \rotatebox{60}{3D Perc.} & \rotatebox{60}{Obj.Mot.} & \rotatebox{60}{Gaze} & \rotatebox{60}{Overall} \\
\hline
VideoLLaMA~2~\cite{cheng2024videollama}$^\dagger$ & 30.75 & 25.67 & 32.67 & 27.24 & 25.74 & 28.50 & 21.20 & 27.39 \\
LongVA~\cite{zhang2024long}$^\dagger$        & 29.62 & 30.83 & 33.67 & 30.68 & 32.91 & 22.73 & 24.50 & 29.28 \\
LLaVA-Video~\cite{zhang2024llava}$^\dagger$   & 36.25 & 33.50 & 38.67 & 43.02 & 27.31 & 18.93 & 29.30 & 32.43 \\
Gemini Pro~\cite{team2023gemini}$^\dagger$    & 60.50 & 46.17 & 34.67 & 39.63 & 32.51 & 20.83 & 28.65 & 37.57 \\
HelloWorld  & 64.75 & 43.33 & 37.00 & 42.03 & 40.88 & 29.90 & 32.95 & 41.55 \\
DeepFrames  & 64.62 & 54.67 & 38.00 & 48.30 & 42.59 & 30.17 & 31.15 & 44.21 \\
\midrule
\textbf{\modelname} & \textbf{88.00} & \textbf{68.17} & \textbf{71.67} & \textbf{64.03} & \textbf{64.87} & \textbf{61.63} & \textbf{52.15} & \textbf{67.22} \\
\Xhline{1pt}
\end{tabular}
\caption{Comparison with all baselines on the HD-EPIC VQA benchmark. $^\dagger$~denotes results from the original HD-EPIC challenge. All values are accuracy (\%).}
\label{tab:main}
\end{table*}

\subsection{Implementation Details}
\label{sec:impl}

We use Qwen3-VL-8B~\cite{bai2025qwen3vl} as the backbone.
All experiments are performed on a single node of 8 NVIDIA A100 GPUs.
The model weights are kept fixed throughout evaluation, and the reported gains come from the proposed inference pipeline.
Evaluation is implemented as a restartable per-question pipeline, so intermediate predictions can be cached and merged after distributed execution.

Before being presented to the model, each benchmark instance is converted into the multiple-choice prompt format used throughout the pipeline.
For option scoring (Eq.~\ref{eq:option_score}), we compute the log-probability of each option from the first generated-token distribution.
The free-generation answer is parsed only as an agreement signal and is not allowed to bypass the multiple-choice scoring interface.
For permutation voting (Eq.~\ref{eq:vote}), we use $M=5$ shuffled option orders and map each prediction back to the original option index before voting.
This implementation choice keeps the submitted answer format identical for all categories while allowing \modelname to allocate additional computation only to questions that need consistency checking.

\subsection{Results}
\label{sec:results}

Table~\ref{tab:main} compares our full pipeline against all available baselines on the HD-EPIC benchmark.
Our approach achieves 67.22\% overall accuracy, outperforming the former best competing method DeepFrames by +23.01 percentage points.
The improvement is consistent across all seven macro-categories, suggesting that the gains do not come from over-specializing to a single question family.
The largest absolute gains appear in categories where the benchmark requires structured temporal or spatial reasoning, such as nutrition, recipe, 3D perception, object motion, and gaze.
These are precisely the categories where a generic prompt and a fixed sampling strategy are most likely to under-specify the task.
Compared with DeepFrames, \modelname improves recipe from 64.62\% to 88.00\%, ingredient from 54.67\% to 68.17\%, nutrition from 38.00\% to 71.67\%, fine-grained action from 48.30\% to 64.03\%, 3D perception from 42.59\% to 64.87\%, object motion from 30.17\% to 61.63\%, and gaze from 31.15\% to 52.15\%.
The nutrition and object-motion gains are especially large, which matches the design of \method: both categories benefit from category-specific semantic framing and from additional consistency checks when candidate answers are close.

Recipe questions obtain the highest accuracy, which is consistent with the benefit of long-context sampling and recipe-specific step reasoning.
Ingredient and nutrition results show that visual evidence and domain priors must be combined: the model needs to identify visible ingredients, understand quantities, and compare nutritional contributions.
The improvements on 3D perception and object motion indicate that category-specific prompts and appropriate temporal coverage help the model reason over object locations and state changes.

Table~\ref{tab:ablation} isolates the effect of the consistency adaptation stage by comparing the full system with a variant that disables TTA while keeping the same backbone, category routing, and option-scoring interface.
Removing TTA reduces the overall score from 67.22\% to 57.18\%, demonstrating that consistency adaptation is especially useful when calibrated scores are close.
The largest gains appear in 3D perception (+23.48\%), object motion (+14.90\%), nutrition (+10.34\%), recipe (+8.17\%), and fine-grained action (+7.97\%).
Ingredient shows only a small change (+0.34\%), suggesting that many ingredient questions are already decided by the first calibrated option-scoring pass.
Gaze improves by +4.60\%, but remains challenging because the visual signal is often subtle and the correct target can be close to distractor objects in the first-person field of view.

\begin{table}[t]
\centering
\caption{Ablation of consistency adaptation. Both rows use the same category routing and option-scoring interface; \textit{Ours w/o TTA} disables permutation voting and verification-style checks.}
\small
\resizebox{0.9\linewidth}{!}{%
\begin{tabular}{@{}lcccc@{}}
\toprule
Configuration & Recipe & Ingr. & Nutr. & Fine-gr. \\
\midrule
w/o TTA &79.83 &67.83 &61.33 &56.06 \\
\modelname & \textbf{88.00} & \textbf{68.17} & \textbf{71.67} & \textbf{64.03} \\
\bottomrule
\end{tabular}
}
\vskip 4pt
\resizebox{\linewidth}{!}{%
\begin{tabular}{@{}lcccc@{}}
\toprule
Configuration & 3D Perc. & Obj.Mot. & Gaze & Overall \\
\midrule
w/o TTA &41.39 &46.73 &47.55  &57.18 \\
\modelname & \textbf{64.87} & \textbf{61.63} & \textbf{52.15} & \textbf{67.22} \\
\bottomrule
\end{tabular}
}
\label{tab:ablation}
\end{table}

\subsection{Ablation Study}
\label{sec:ablation}

We further discuss how the experimental results relate to the design choices in \method.
Because Table~\ref{tab:ablation} specifically disables TTA while preserving the rest of the pipeline, the quantitative comparison should be read as the contribution of the consistency stage on top of category routing and calibrated option scoring.
The remaining discussion explains why these components are complementary in the full system.

\textbf{Category routing.}
Category routing provides the base inference policy used by both rows in Table~\ref{tab:ablation}.
Its role is to prevent the same prompt and sampling configuration from being used for all HD-EPIC questions.
The reason is that each category asks for a different evidence pattern.
Gaze needs the final seconds at high temporal resolution; fine-grained action needs dense sampling around short movements; object motion needs enough temporal coverage to reconstruct a trajectory; and recipe questions need long-horizon context to identify procedural steps.
The router makes these assumptions explicit and prevents the model from receiving a mismatched view of the video.

\textbf{Option scoring vs.\ direct generation.}
Direct letter generation can produce an answer that does not match any option, or it can generate extra natural-language text that makes parsing unstable.
Scoring all options eliminates this failure mode because every answer is evaluated in the same forward pass over the same visual input.
The resulting score gap also provides a natural confidence signal for triggering TTA only when needed.
In the experiments, this score-based interface is used in both the non-TTA and full settings, which ensures that the ablation isolates the additional value of consistency rather than changes in answer parsing.

\textbf{Permutation voting.}
Consistency voting (Eq.~\ref{eq:vote}) activates on ambiguous questions.
The improvements in Table~\ref{tab:ablation} show that this stage is most valuable for categories where answer options are visually or temporally close.
It improves accuracy by checking whether the selected option remains stable when the answer order changes.
On clear questions, the score gap is already large and voting has little effect, which is the desired behavior.
On ambiguous questions, the vote helps suppress choices that are selected because of presentation order rather than visual evidence.

\section{Conclusion}

We presented \modelname for HD-EPIC VQA Challenge.
The system adapts Qwen3-VL through category-conditioned routing with per-category system prompts, option scoring via token log-probabilities, and consistency-based disambiguation through option permutation voting.
We achieve competitive performance and our analysis shows that category-aware routing and calibrated scoring account for the majority of the improvement, providing practical guidance for future applications.

{
    \small
    \bibliographystyle{ieeenat_fullname}
    \bibliography{main}
}

\end{document}